\def\year{2017}\relax
\documentclass[letterpaper]{article}
\usepackage{aaai17}
\usepackage{times}
\usepackage{helvet}
\usepackage{courier}

\usepackage[normalem]{ulem}
\usepackage{xspace}
\usepackage{amsmath}
\usepackage{amssymb}
\usepackage{url}
\usepackage[draft]{todonotes}   
\usepackage{quoting}
\quotingsetup{vskip=3pt,leftmargin=5mm,rightmargin=5mm}

\usepackage{xcolor}

\makeatletter
\def\uwave{\bgroup \markoverwith{\lower3.5\p@\hbox{\sixly \textcolor{red}{\char58}}}\ULon}
\font\sixly=lasy6 
\makeatother

\newcommand{\jlm}{scRNN\xspace}

\newcommand{\ms}{\text{Commercial A}\xspace}
\newcommand{\gr}{\text{Commercial B}\xspace}

\newcommand{\cnn}{\text{CharCNN}\xspace}

\newcommand{\cambridge}{{\em Cmabrigde Uinervtisy}\xspace}

\newcommand{\newcite}[1]{\citeauthor{#1} (\citeyear{#1})}

\newcommand{\clsp}{\ensuremath{{}^\dagger}}
\newcommand{\hltcoe}{\ensuremath{{}^\ddagger}}

\begin{document}
%
\title{Robsut Wrod Reocginiton via Semi-Character Recurrent Neural Network\thanks{Parts of this manuscript are intentionally jumbled to demonstrate the robust word processing ability of you, the reader.}}

\author{
 Keisuke Sakaguchi\clsp
  \ \ \ \ Kevin Duh\hltcoe 
  \ \ \ \ Matt Post\hltcoe 
  \ \ \ \ Benjamin Van Durme\clsp\hltcoe \\
  \clsp Center for Language and Speech Processing, Johns Hopkins University \\
  \hltcoe Human Language Technology Center of Excellence, Johns Hopkins University \\
 {\tt \{keisuke,kevinduh,post,vandurme\}@cs.jhu.edu}
}

\maketitle


\begin{abstract}
Language processing mechanism by humans is generally more robust than computers. 
The \cambridge ({\em Cambridge University}) effect from the psycholinguistics literature has demonstrated such a robust word processing mechanism, where jumbled words (e.g. Cmabrigde / Cambridge) are recognized with little cost.
On the other hand, computational models for word recognition (e.g. spelling checkers) perform poorly on data with such noise.

Inspired by the findings from the \cambridge effect, we propose a word recognition model based on a semi-character level recurrent neural network (\jlm).
In our experiments, we demonstrate that \jlm has significantly more robust performance in word spelling correction (i.e. word recognition) compared to existing spelling checkers and character-based convolutional neural network.
Furthermore, we demonstrate that the model is cognitively plausible by replicating a psycholinguistics experiment about human reading difficulty using our model.
\end{abstract}


\section{Introduction}
\label{sect:intro}
Despite the rapid improvement in natural language processing by computers, humans still have advantages in situations where the text contains noise. 
For example, the following sentences, introduced by a psycholinguist \cite{davis2003aoccdrnig}, provide a great demonstration of the robust word recognition mechanism in humans. 
\begin{quoting}
{\em Aoccdrnig to a rscheearch at Cmabrigde Uinervtisy, it deosn't mttaer in waht oredr the ltteers in a wrod are, the olny iprmoetnt tihng is taht the frist and lsat ltteer be at the rghit pclae. The rset can be a toatl mses and you can sitll raed it wouthit porbelm. Tihs is bcuseae the huamn mnid deos not raed ervey lteter by istlef, but the wrod as a wlohe.}
\end{quoting}

This example shows the \cambridge ({\em Cambridge University}) effect, which demonstrates that human reading is resilient to (particularly internal) letter transposition.

\begin{figure}[t]
	\centering
	\includegraphics[width=80mm]{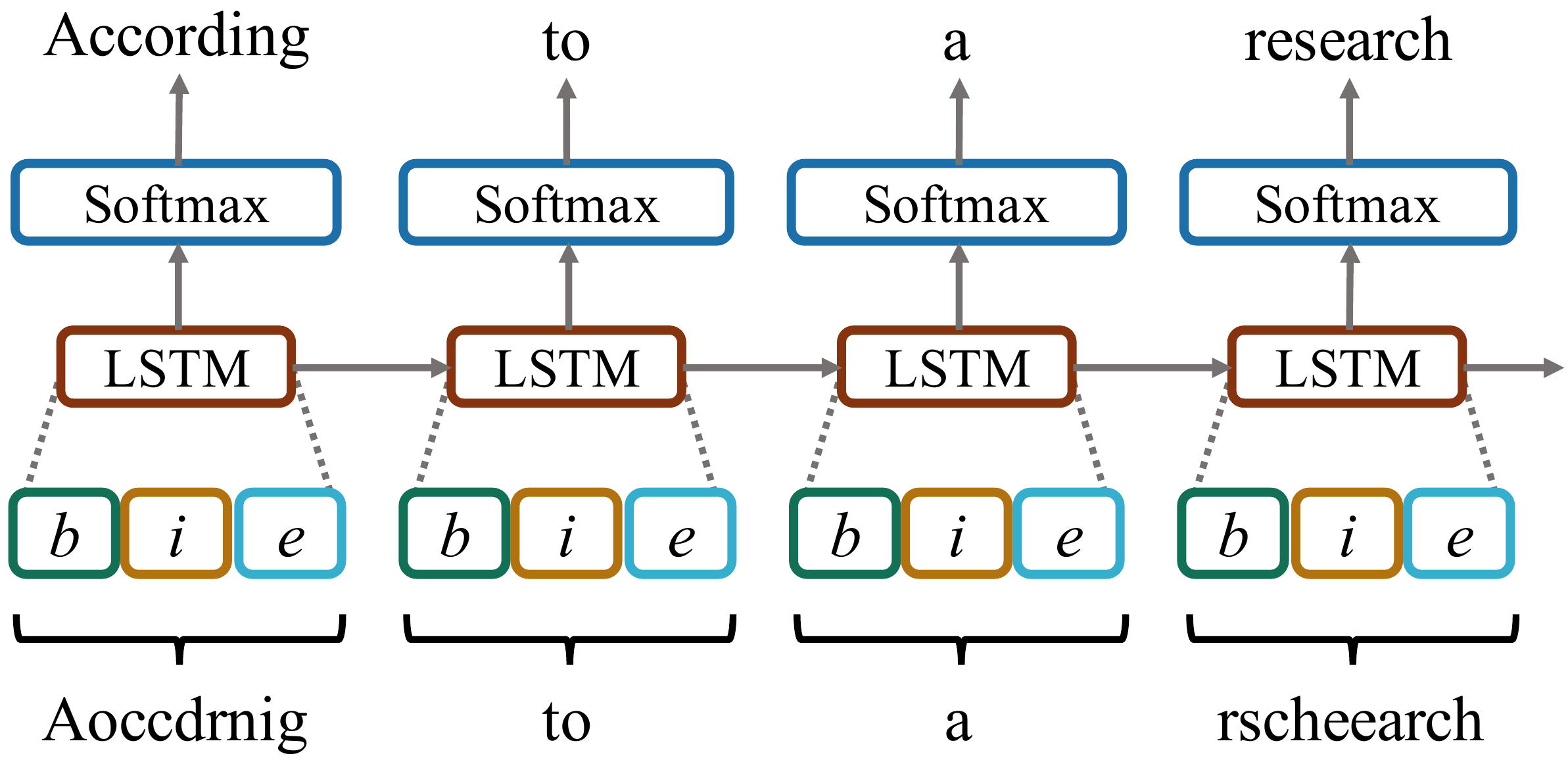}
	\caption{\label{fig:scrnnlm} Schematic Illustration of semi-character recurrent neural network (\jlm).}
\end{figure}

Robustness is an important and useful property for various tasks in natural language processing, and we propose a computational model which replicates this robust word recognition mechanism.
The model is based on a standard recurrent neural network (RNN) with a memory cell as in long short-term memory \cite{hochreiter1997long}.
We use an RNN because it has shown to be state-of-the-art language modeling \cite{DBLP:conf/interspeech/MikolovKBCK10}
and it is also flexible to realize the findings from the \cambridge effect.
Technically, the input layer of our model consists of three sub-vectors: beginning ($b$), internal ($i$), and ending ($e$) character(s) of the input word (Figure \ref{fig:scrnnlm}).
This semi-character level recurrent neural network is referred as \jlm.

\begin{table*}[t]
\centering
\begin{tabular}{|lcccc|} \hline
Cond.  & Example & \# of fixations & Regression(\%) & Avg. Fixation (ms) \\ \hline
  N    & \small{The boy could not solve the problem so he asked for help.} & 10.4$^{\ \ }$ & 15.0$^{\ \ }$  & 236$^{\ \ }$  \\ 
INT  & \small{The boy cuold not slove the probelm so he aksed for help.} & 11.4$^{*}$ & 17.6$^{*}$  & 244$^{*}$  \\
END  & \small{The boy coudl not solev the problme so he askde for help.} & 12.6$^{\dagger}$ & 17.5$^{*}$  & 246$^{*}$  \\
BEG  & \small{The boy oculd not oslve the rpoblem so he saked for help.} & 13.0$^{\ddagger}$ & 21.5$^{\dagger}$  & 259$^{\dagger}$  \\ \hline
\end{tabular}
  \caption{Example sentences and results for measures of fixation excerpt from \newcite{Rayner01032006}. There are 4 conditions: N = normal text; INT = internally jumbled letters; END = letters at word endings are jumbled; BEG = letters at word beginnings are jumbled. Entries with $^*$ have statistically significant difference from the condition N ($p < 0.01$) and those with $^{\dagger}$ and $^{\ddagger}$ differ from $^*$ and $^{\dagger}$ with $p < 0.01$ respectively.}
\label{tab:previous}
\end{table*}

First, we review previous work on the robust word recognition mechanism from psycholinguistics literature.
Next, we describe technical details of \jlm which capture the robust human mechanism using 
recent developments in neural networks. 
As closely related work, we explain character-based convolutional neural network (\cnn) proposed by \newcite{DBLP:journals/corr/KimJSR15}.
Our experiments show that the \jlm significantly outperforms commonly used spelling checkers and \cnn by (at least) 42\% for jumbled word correction and 3\% and 14\% in other noise types (insertion and deletion).
We also show that \jlm replicates recent findings from psycholinguistics experiments on reading difficulty depending on the position of jumbled letters, which indicates that \jlm successfully mimics (at least a part of) the robust word recognition mechanism by humans.


\section{Raeding Wrods with Jumbled Lettres}
\label{sect:previous}
\begin{figure}[t]
	\centering
	\includegraphics[width=65mm]{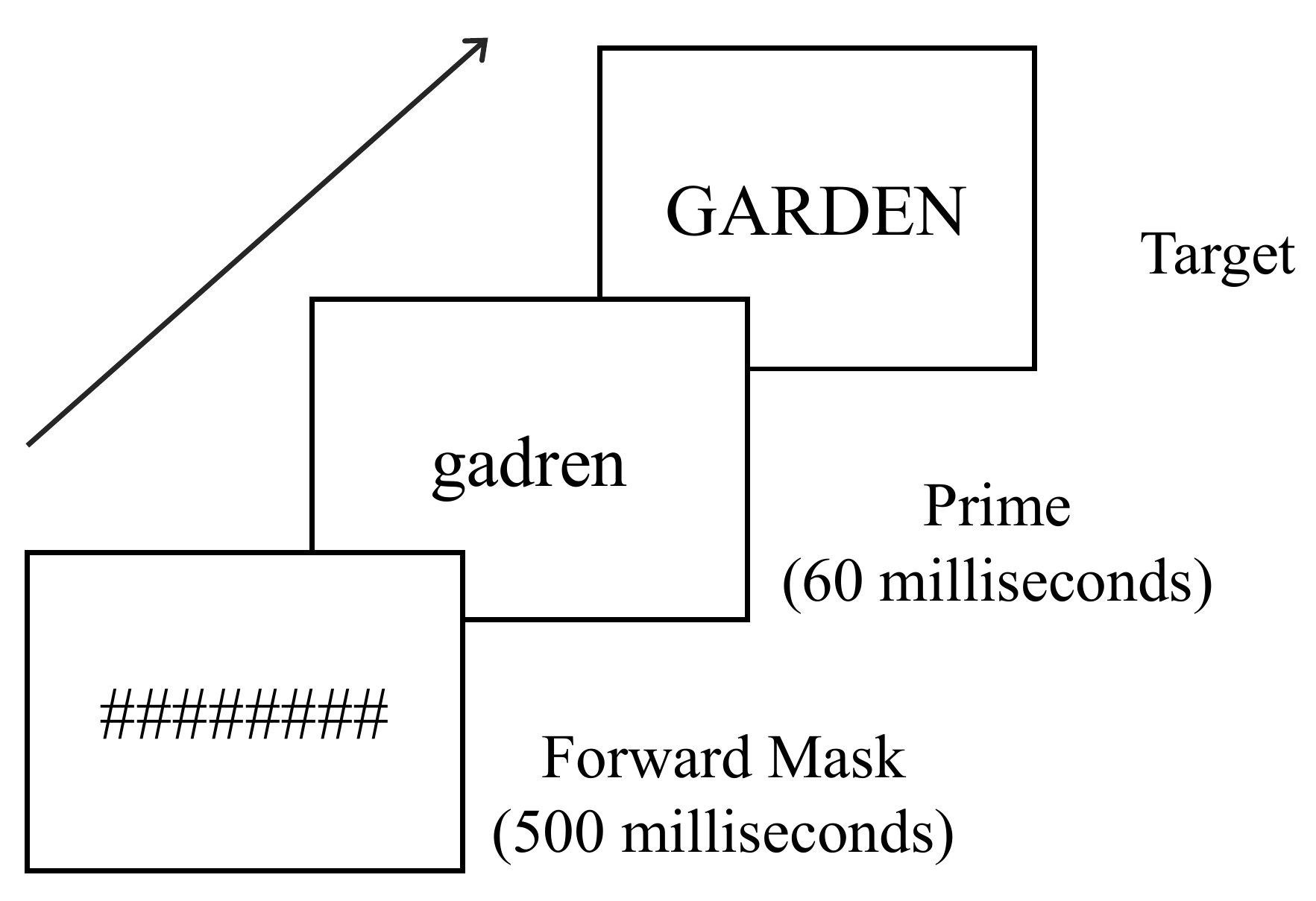}
	\caption{\label{fig:priming} Example of the masked priming procedure.}
\end{figure}

Sentence processing with jumbled words has been a major research topic in psycholinguistics literature.
One popular experimental paradigm is {\em masked priming}, in which a (lower-cased) stimulus, called {\em prime}, is presented for a short duration (e.g. 60 milliseconds) followed by the (upper-cased) target word, and participants are asked to judge whether the target word exists in English as quickly as possible (Figure \ref{fig:priming}).\footnote{There is another variant for masked priming technique, where backward mask is inserted between the prime and target in addition to the forward mask.}
The prime is consciously imperceptible due to the instantaneous presentation but it proceeds to visual word recognition by participants.
The masked priming paradigm allows us to investigate the machinery of lexical processing and the effect of prime in a pure manner.

\newcite{forster1987masked} show that a jumbled word (e.g. gadren-GARDEN) facilitates primes as large as identity primes (garden-GARDEN) and these results have been confirmed in cases where the transposed letters are not adjacent (caniso-CASINO) \cite{Perea2004231} and even more extreme cases (sdiwelak-SIDEWALK) \cite{doi:10.1080/01690960701579722}.

These findings about robust word processing mechanism by humans have been further investigated by looking at other types of noise in addition to simple letter transpositions.
\newcite{Humphreys1990517} show that deleting a letter in a word still produces significant priming effect (e.g. blck-BLACK), and similar results have been shown in other research \cite{Peressotti1999,grainger2006letter}.
\newcite{van2006study} demonstrate that a priming effect remains when inserting a character into a word (e.g. juastice-JUSTICE).

Another popular experimental paradigm in psycholinguistics is {\em eye-movement} tracking.
In comparison to the masked priming technique, eye-movement paradigm provides data from normal reading process by participants.
Regarding word recognition, the eye-tracking method has shown the relationship between a word difficulty and the eye fixation time on the word: when a word is difficult to process, average time to fixation becomes long.
In addition, words that are difficult to process often induce regressions to words previously read.

With the eye-movement paradigm, 
\newcite{Rayner01032006} and \newcite{johnson2007transposed} conduct detailed experiments on the robust word recognition mechanism with jumbled letters.
They show that letter transposition affects fixation time measures during reading depending on which part of the word is jumbled. 
Table \ref{tab:previous} presents the result from \newcite{Rayner01032006}. 
It is obvious that people can read smoothly (i.e. smaller number of fixations, regression, and average of fixation duration) when a given sentence has no noise (referring to this condition as N).
When the characters at the beginning of words are jumbled (referring to this condition as BEG), participants have more difficulty (e.g. longer fixation time).
The other two conditions, where words are internally jumbled (INT) or letters at word endings are jumbled (END), have similar amount of effect, although the number of fixations between them showed a statistically significant difference ($p < 0.01$).
In short, the reading difficulty with different jumble conditions is summarized as follows: N $<$ INT $\leq$ END $<$ BEG. 

It may be surprising that there is statistically significant difference between END and BEG conditions despite the difference being very subtle (i.e. fixing either the first or the last character).
This result demonstrates the importance of beginning letters for human word recognition.\footnote{While there is still ongoing debate in the psycholinguistics community as to exactly how (little) the order of internal letters matter, here we follow the formulation of \newcite{Rayner01032006}, considering only the letter order distinctions of BEG, INT, and END.}


\section{Semi-Character Recurrent Neural Network}
\label{sect:jlm}
In order to achieve the human-like robust word processing mechanism, we propose a semi-character based recurrent neural network (\jlm).
The model takes a semi-character vector ($x$) for a given jumbled word, and predicts a (correctly spelled) word ($y$) at each time step.
The structure of \jlm is based on a standard recurrent neural network, where current input ($x$) and previous information is connected through hidden states ($h$) by applying a certain (e.g. sigmoid) function ($g$) with linear transformation parameters ($W$) and the bias ($b$) at each time step ($t$).

One critical issue of vanilla recurrent neural networks is that it is unable to learn long distance dependency in the inputs due to the vanishing gradient \cite{bengio1994learning}.
To address the problem, \newcite{hochreiter1997long} introduced long short-term memory (LSTM), which is able to learn long-term dependencies by adding a memory cell ($c$).
The memory cell has an ability to discard or keep previous information in its state.
Technically, the LSTM architecture is given by the following equations, 
\begin{align}
  i_{n}&=\sigma\left( W_i\left[ h_{n-1}, x_{n}\right] +b_i\right) \\
  f_{n}&=\sigma\left( W_f\left[ h_{n-1}, x_{n}\right] +b_f\right) \\
  o_{n}&=\sigma\left( W_o\left[ h_{n-1}, x_{n}\right] +b_o\right) \\
  g_{n}&=\sigma\left( W_g\left[ h_{n-1}, x_{n}\right] +b_g\right) \\
  c_{n}&=f_n\odot c_{n-1}+i_{n}\odot g_n \\
  h_{n}&=o_{n}\odot \tanh \left( c_{n}\right)
\end{align}
where $\sigma$ is the (element-wise) sigmoid function and $\odot$ is the element-wise multiplication.

While a standard input vector for RNN derives from either a word or a character, the input vector in \jlm consists of three sub-vectors ($b_n, i_n, e_n$) that correspond to the characters' position.
\begin{align}
  x_n &= \begin{bmatrix}
           b_{n} \\
           i_{n} \\
           e_{n}
        \end{bmatrix}
\end{align}
The first and third sub-vectors ($b_{n}$, $e_{n}$) represent the first and last 
character of the $n$-th word. 
These two sub-vectors are therefore one-hot representations.
The second sub-vector ($i_{n}$) represents a bag of characters of the word without the initial and final positions. 
For example, the word ``University'' is represented as $b_n = \{U=1\}$, $e_n = \{y=1\}$, and $i_n = \{e=1, i=2, n=1, s=1, r=1, t=1, v=1\}$, with all the other elements being zero. 
The size of sub-vectors ($b_n, i_n, e_n$) is equal to the number of characters ($N$) in our language, and $x_n$ has therefore the size of $3N$ by concatenating the sub-vectors.

Regarding the final output (i.e. predicted word $y_n$), the hidden state vector ($h_n$) of the LSTM is taken as input to the following softmax function layer with a fixed vocabulary size ($v$).
\begin{align}
  y_{n}&=\dfrac {\exp \left( W_h \cdot h_{n}\right) }{\sum _{v}\exp \left( W_h \cdot h_{n}\right) }
\end{align}
We use the cross-entropy training criterion applied to the output layer as in most LSTM language modeling works; the model learns the weight matrices ($W$) to maximize the likelihood of the training data. 
This should approximately correlate with maximizing the number of exact word match in the predicted outputs. 
Figure \ref{fig:scrnnlm} shows a pictorial overview of \jlm.

In order to check if the \jlm can recognize the jumbled words correctly, we test it in spelling correction experiments.
If the hypothesis about the robust word processing mechanism is correct, \jlm will also be able to read sentences with jumbled words robustly.


\section{Character-based Neural Network}
\label{sect:cnn}
Another possible approach to deal with reading jumbled words by neural networks is (pure) character-level neural network \cite{sutskever2011generating}, where both input and output are characters instead of words.
The character-based neural networks have been investigated and used for a variety of NLP tasks such as segmentation \cite{DBLP:journals/corr/Chrupala13}, dependency parsing \cite{ballesteros-dyer-smith:2015:EMNLP}, machine translation \cite{DBLP:journals/corr/LingTDB15}, and text normalization  \cite{chrupala:2014:P14-2}.

For spelling correction, \newcite{schmaltz-EtAl:2016:BEA11} uses character-level convolutional neural networks (\cnn) proposed by \newcite{DBLP:journals/corr/KimJSR15}, in which the input is character but the prediction is at the word-level.
More technically, according to \newcite{DBLP:journals/corr/KimJSR15}, \cnn concatenates the character embedding vectors into a matrix $P_n \in \mathbb{R}^{d \times l}$ whose $k$-th column corresponds to the $k$-th character embedding vector (size of $d$) of $n$-th word which contains $l$ characters.
A narrow convolution is applied between $P$ and {\em filter} $H \in \mathbb{R}^{d \times w}$ of width $w$, and then {\em feature map} $f_n\in\mathbb{R}^{l-w+1}$ is obtained by the following transformation\footnote{In the equation, $P_n[:, k:k+w-1]$ means the $k$-to-($k+w-1$)-th column of $P_n$.} with a bias $b$.
\begin{align}
  f_n = \tanh(\text{Tr}(P_n[:, k:k+w-1] H^{T}) + b)
\end{align}
This is interpreted as a process of capturing important feature $f$  with filter $H$ to maximize the predicted word representation $y_n$ by the {\em max-over-time}:
\begin{align}
y_n = \max_k f_n[k]
\end{align}

Although \cnn and \jlm have some similarity in terms of using a recurrent neural network, \cnn is able to store richer representation than \jlm. 
In the following section, we compare the performance of \cnn and \jlm with respect to jumbled word recognition task.


\section{Experiments}
\label{sect:exp}
We conducted spelling correction experiments to judge how well \jlm can recognize noisy word sentences. 
In order to make the task more realistic, we tested three different noise types: {\it jumble}, {\it delete}, and {\it insert}, 
where the {\it jumble} changes the internal characters (e.g. Cambridge $\to$ Cmbarigde), 
{\it delete} randomly deletes one of the internal characters (Cambridge $\to$ Camridge),
and {\it insert} randomly inserts an alphabet into an internal position (Cambridge $\to$ Cambpridge).
None of the noise types change the first and last characters.
We used Penn Treebank for training, tuning, and testing.\footnote{Section 2-21 for training, 22 for tuning, and 23 for test \url{https://catalog.ldc.upenn.edu/ldc99t42}.
The data includes 39,832 sentences in training set (898k/950k tokens are covered by the top 10k vocabulary), 1,700 sentences in the tuning set (coverage 38k/40k), and 2,416 sentences in test set (coverage 54k/56k).}

\begin{table*}[t]
\centering
\small
\begin{tabular}{|p{2cm}|p{13cm}|}\hline
Original & Aoccdrnig to a rscheearch at Cmabrigde Uinervtisy , it deos n't mttaer in waht oredr the ltteers in a wrod are , the olny iprmoetnt tihng is taht the frist and lsat ltteer be at the rghit pclae . The rset can be a toatl mses and you can sitll raed it wouthit porbelm . Tihs is bcuseae the huamn mnid deos not raed ervey lteter by istlef , but the wrod as a wlohe .   \\ \hline
Correct  & According to a researcher at Cambridge University , it does n't matter in what order the letters in a word are , the only important thing is that the first and last letter be at the right place . The rest can be a total mess and you can still read it without problem . This is because the human mind does not read every letter by itself ,  but the word as a whole . \\ \hline\hline
\cnn \ \ \ \ \ \ \ \ \ \ \ \ \ \ (\citeauthor{DBLP:journals/corr/KimJSR15}) & According to a \uwave{research} at Cambridge \uwave{Minority} , it \uwave{deck} n’t \uwave{mother} in \uwave{wait} or the letters in a \uwave{wood} are , the \uwave{tony} \uwave{Vermont} \uwave{timing} is \uwave{taxi} the \uwave{tourist} and \uwave{sat} letter be at the \uwave{fruit} \uwave{pile} . The \uwave{reset} can be a total \uwave{uses} and you can \uwave{vital} \uwave{rake} it \uwave{worthy} \uwave{parallel} . \uwave{Mips} is \uwave{abuse} the human \uwave{trim} \uwave{deck} not \uwave{rake} \uwave{survey} \uwave{latter} by \uwave{leftist} , but the \uwave{wood} as a whole . \\ \hline
Enchant & \uwave{Ecuadoran} to a \uwave{searcher} at \uwave{Brigade} \uwave{Nerviness} , it does n't matter in what order the letters in a word are , the only \uwave{omnipresent} thing is that the \uwave{freest} and \uwave{slat} letter be at the right place . The rest can be a total mess and you can still read it \uwave{outhit} \uwave{corbel} . \uwave{Tish} is \uwave{Ceausescu} , the human mind does not read \uwave{Hervey} letter by \uwave{leftist} , but the word as a whole . \\ \hline
\ms       & \uwave{Occurring} to a \uwave{scholarch} at Cambridge \uwave{Inertias} , it does n't matter in what order the letters in a word are , the only impotent thing is that the first and last letter be at the right place . The rest can be a total mess and you can still read it \uwave{outhit} problem . This is \uwave{bcuseae} the human mind does not read every letter by \uwave{istle} , but the word as a whole . \\ \hline
\gr       & \uwave{Aoccdrnig} to a  \uwave{rscheearch} at \uwave{Cmabrigde} \uwave{Uinervtisy} , it does n't matter in what order the letters in a word are , the only \uwave{iprmoetnt} thing is that the first and last letter be at the right place . The rest can be a total mess and you can still read it \uwave{wouthit} problem . \uwave{Tihs} is \uwave{bcuseae} the human mind does not read every letter by itself , but the word as a whole . \\ \hline
\jlm \ \ \ \ \ \ \ \ (proposed)  & According to a \uwave{research} at Cambridge University , it does n't matter in what order the letters in a word are , the only important thing is that the first and last letter be at the right place . The rest can be a total mess and you can still read it without problem . This is because the human mind does not read every letter by itself , but the word as a whole . \\ \hline
\end{tabular}
\caption{Example spelling correction outputs for the \cambridge sentences. Words that the system failed to correct are \uwave{underlined}.
\cnn stands for the character-based convolutional neural network by \newcite{DBLP:journals/corr/KimJSR15}.}
\label{tab:cambridge}
\end{table*}

The input layer of \jlm consists of a vector with length of 76 (A-Z, a-z and 24 symbol characters).
The hidden layer units had size 650,
and total vocabulary size was set to 10k.
We applied one type of noise to every word, but words with numbers (e.g. 1980s) and short words (length $\leq 3$) were not subjected to jumbling, and therefore these words were excluded in evaluation.
We trained the model by running 5 epochs with (mini) batch size 20.
We set the backpropagation through time (BPTT) parameter to 3: \jlm updates weights for previous two words ($x_{n-2}, x_{n-1}$) and the current word ($x_n$).

For comparison, we evaluated \cnn on the same training data\footnote{For \cnn, we employed the codebase available at \url{https://github.com/yoonkim/lstm-char-cnn.git}}, and also compared widely-used spelling checkers (Enchant\footnote{\url{http://www.abisource.com/projects/enchant/}}, \ms, and \gr \footnote{We anonymized the name of the commercial product.}).

\begin{table}[t]
\centering
\begin{tabular}{|l|c|c|c|} \hline
         & Jumble  & Delete & Insert \\ \hline
\cnn (\citeauthor{DBLP:journals/corr/KimJSR15})  & 17.17   & 21.30  & 35.00  \\
Enchant  & 57.15   & 37.01  & 88.54  \\
\jlm (proposed)    & {\bf 98.96}   & {\bf 85.74}  & {\bf 96.70}  \\ \hline
\end{tabular}
\caption{Spelling correction accuracy (\%) with different error types on the entire test set.}
\label{tab:spelling}
\end{table}

\begin{table}[t]
\centering
\begin{tabular}{|l|c|c|c|} \hline
         & Jumble  & Delete & Insert \\ \hline
\cnn (\citeauthor{DBLP:journals/corr/KimJSR15})  & 16.18   & 19.76  & 35.53  \\
Enchant  & 57.59   & 35.37  & 89.63  \\
\ms      & 54.81   & 60.19  & 93.52  \\
\gr      & 54.26   & 71.67  & 73.52  \\
\jlm (proposed)    & {\bf 99.44}   & {\bf 85.56}  & {\bf 97.04}  \\ \hline
\end{tabular}
\caption{Spelling correction accuracy (\%) with different error types on the subset of test set (50 sentences).}
\label{tab:spelling_small}
\end{table}

\subsection{Spelling correction results}
\label{sect:exp1}

Table \ref{tab:cambridge} presents example outputs for the \cambridge sentence by each model.\footnote{The \cambridge sentences contains jumbling as well as deletion, insertion, and replacement of characters. Note that we used a single \jlm (Jumble), and didn't train \jlm separately for each error type in this example.}
It may be surprising that \cnn performs poorly compared with other spelling checkers.
This is probably because the \cnn highly depends on the order of characters in the word and the transposed characters adversely affected the recognition performance.
Enchant, \ms, and \gr tend to fail long word correction. 
This may be because these models are not designed for severely jumbled input but they are likely to depend on edit distance between the incorrect and correct words.
While these existing models struggle with correcting \cambridge sentence, we see that \jlm demonstrates significantly better recognition ability.
The only error in \jlm may be because the last character (rscheearc{\bf h}) activated the \jlm nodes strongly toward {\em research} instead of {\em researcher}.\footnote{There is also an deletion of 'r'.}

\begin{figure}[t]
	\centering
	\includegraphics[width=85mm]{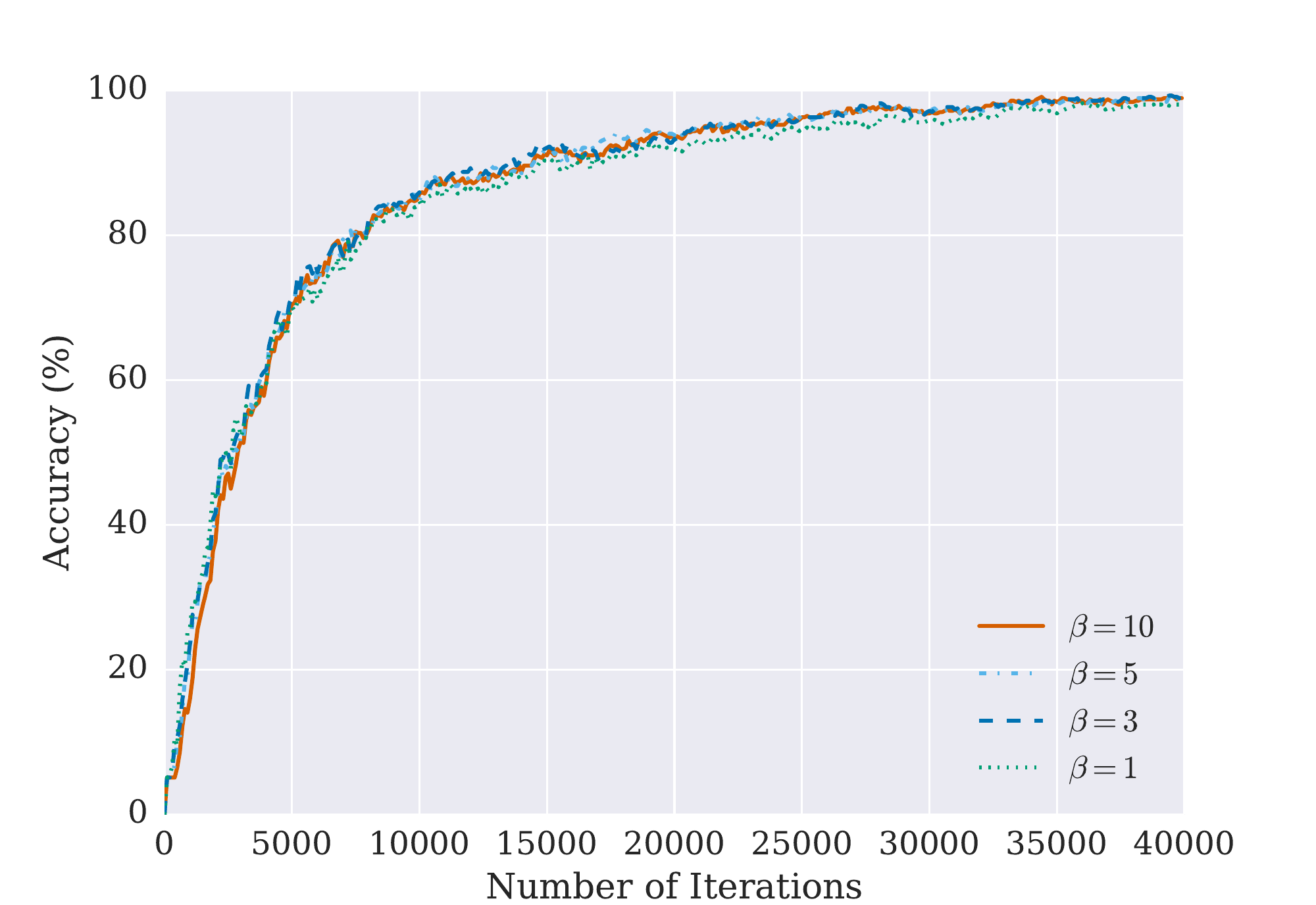}
	\includegraphics[width=85mm]{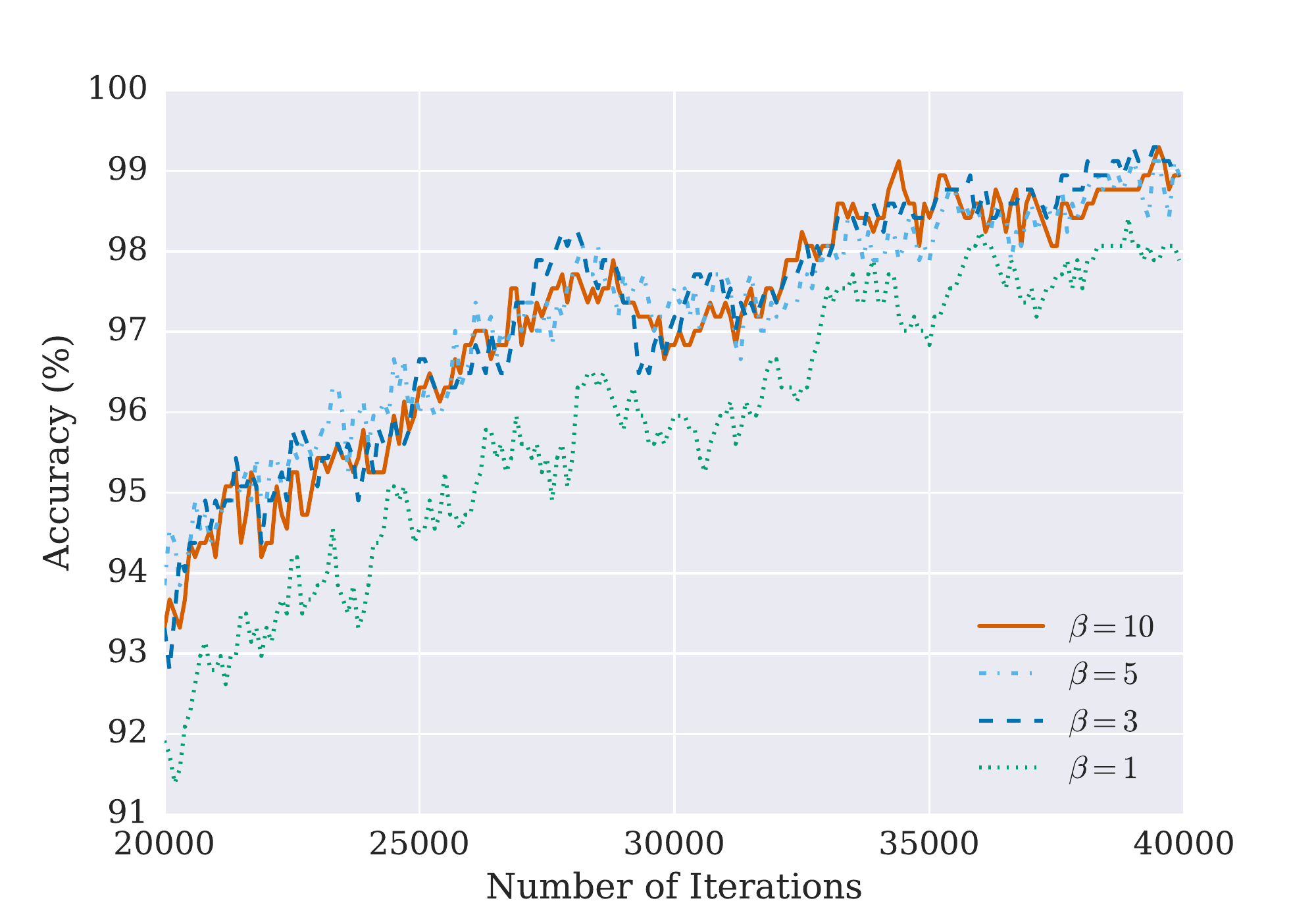}
	\caption{\label{fig:learning} Learning curve of training \jlm with different BPTT parameter (on dev set): first 40k iterations (top) and its enlarged view between 20k and 40k iterations (bottom). }
\end{figure}

Table \ref{tab:spelling} and \ref{tab:spelling_small} show the overall result on the test set with respect to noise type.
We also tested spelling correction on a small subset (50 sentences) because of the API limits etc. of commercial systems.
Overall, as seen in the example above, \jlm significantly outperforms the other spelling checker models across all three noise types.
Since \jlm is especially designed for jumbled word recognition, it is not surprising that it performs particularly well on {\it jumble} noise. 
However, it is striking that \jlm outperforms the other models in deletion and insertion errors as well.\footnote{It is important to note that some commercial systems have constraints of the model size \cite{church-hart-gao:2007:EMNLP-CoNLL2007}.}
This clearly demonstrates the robustness of \jlm.

The relatively large drop in {\it delete} in \jlm may be because the information lost by deleting character is significant.
For example, when the word {\em place} has dropped the character {\em l}, the surface form becomes {\em pace}, which is also a valid word. 
Also, the word {\em mess} with {\em e} being deleted produces the form of {\em mss}, which can be recovered as {\em mess}, {\em mass}, {\em miss}, etc. 
In the \cambridge sentences, in both cases, the local context support other phrase such as `at the right {\em pace/place}' and `a total {\em mass/mess}'.
These examples clearly demonstrate that deleting characters harm the word recognition more significantly than other noise types.
All the models perform relatively well on {\it insert} noise, indicating that adding extraneous information 
by inserting a letter does not change the original information significantly.

With respect to a learning curve on \jlm (Figure \ref{fig:learning}, top), we found that the model achieves 0.9 (in accuracy) at the 15,000-th iteration of the mini batch. 
This can be made within a hour with a CPU machine, which demonstrates simplicity of \jlm compared with \cnn.

\begin{table}[t]
\centering
\begin{tabular}{|r|cc|} \hline
$\beta$ & Accuracy (\%)     & SD \\ \hline
1 & 98.69	& 0.53 \\
3 & 98.96	& 0.45 \\
5 & 98.91	& 0.40 \\
10 & 98.95  & 0.43 \\ \hline
\end{tabular}
\caption{\jlm accuracy (\%) on jumbled word recognition with different BPTT parameters. There were no statistically significant differences among them.}
\label{tab:bptt}
\end{table}

\begin{table}[t]
\centering
\begin{tabular}{|r|ccc|} \hline
Units & Acc (\%)     & SD & Size (KB)\\ \hline
5  & 24.65	& 2.59 & 236 \\
10 & 48.43	& 3.26 & 435 \\
15 & 73.32	& 3.65 & 632 \\
20 & 84.82	& 2.39 & 830 \\
30 & 94.15	& 1.54 & 1,255 \\
40 & 96.90  & 1.26 & 1,670 \\ 
50 & 98.48  & 0.94 & 2,092 \\
60 & 98.39  & 0.81 & 2,514 \\ \hline
\end{tabular}
\caption{\jlm accuracy (\%), the standard deviation, and the size of model file (KB) on jumbled word recognition with respect to the number of units of LSTM.}
\label{tab:units}
\end{table}

\begin{table*}[t]
\centering
\begin{tabular}{|lcc|} \hline
Cond.  & Example & Accuracy \\ \hline
INT & As a relust , the lnik beewetn the fureuts and sctok mretkas rpiped arapt . & 98.96 \ \ \\ 
END & As a rtelus , the lkni betwene the feturus and soctk msatrek rpepid atarp . & 98.68$^{*}$ \\ 
BEG & As a lesurt , the lnik bweteen the utufers and tocsk makrtes pipred arpat . & 98.12$^{\dagger}$ \\ 
ALL & As a strule , the lnik eewtneb the eftusur and okcst msretak ipdepr prtaa . & 96.79$^{\ddagger}$ \\ \hline
\end{tabular}
  \caption{Example sentences and results for spelling correction accuracy by \jlm variants depending on different jumble conditions: INT = internal letters are jumbled; END = letters at word endings are jumbled; BEG = letters at word beginnings are jumbled; ALL = all letters are jumbled. Entries with $^*$ have a difference with marginal significance from the condition INT ($p = 0.07$) and those with $^\dagger$ and $^\ddagger$ differ from $^*$ and $^\dagger$ with $p < 0.01$ respectively.}
\label{tab:jumble}
\end{table*}

\begin{table}[t]
\centering
\begin{tabular}{|p{0.7cm}|p{6.9cm}|}\hline
Cond.  & Examples of errors (correct/wrong) \\ \hline
INT & Once/once, Under/under, Also/also, there/three, form/from, fares/fears, trail/trial, Broad/Board \\ \hline
END & being/begin, quiet/quite, bets/best, stayed/steady, heat/hate, lost/lots + same errors in INT \\ \hline
BEG & Several/reveal, Growth/worth, host/shot, creditors/directors, views/wives + same errors in INT \\ \hline
ALL & Under/trend, center/recent, licensed/declines, stop/tops + same errors in INT, END, \& BEG \\ \hline
\end{tabular}
  \caption{Error analysis of \jlm variants.}
\label{tab:error}
\end{table}

Figure \ref{fig:learning} also shows the effect of BPTT size ($\beta$), and the accuracy on test set is presented in Table \ref{tab:bptt}.
As explained, $\beta$ indicates the context length of updates during training.
It is surprising that the longer contexts ($\beta = 5, 10$) do not necessarily yield better performance.
This is probably because it rarely happens that the context plays an important role on distinguishing ambiguous representation (e.g. anagrams) in \jlm.
If we take closer look at the learning curve (Table \ref{fig:learning}, bottom), however, there is a clear gap in learning efficiency between with and without contexts (i.e. $\beta$=1 vs. the rest).

Finally, we reduced the number of units in the hidden layer to see the model size and performance of \jlm.
Surprisingly, as Table \ref{tab:units} presents, \jlm with 50 units already achieves comparable results to 650 units (Table \ref{tab:spelling}).
The result suggests that 50 units (2 MB) are enough to distinguish 10k English words.

\subsection{Corroboration with psycholinguistic experiments}
\label{sect:exp2}

As seen in the literature review in psycholinguistics, the position of jumbled characters affects the cognitive load of human word recognition.
We investigate this phenomenon with \jlm by manipulating the structure of input vector.
We replicate the experimental paradigm in \newcite{Rayner01032006}, but using \jlm rather than human subjects. 
We trained \jlm variants depending on different jumble conditions: INT, END, BEG, and ALL. 
INT is the same model as explained in the previous section ($x_n = [b_n, i_n, e_n]^T$).
END represents an input word as a concatenation of the initial character vector ($b$) and a vector for the rest of characters ($x_n = [b_n, i_n + e_n]^T$).
In other words, in END model, the internal and last characters are subject to jumbling. 
BEG model combines a vector for the final character ($e$) and a vector for the rest of characters ($x_n = [b_n + i_n, e_n]^T$). 
In other words, initial and internal characters are subject to jumbling.
In ALL model, all the letters are subject to jumble (e.g. research vs. eesrhrca) and represented as a single vector ($x_n = [b_n + i_n + e_n]^T$).
This is exactly the same as bag of characters.
We trained all the \jlm variants with $\beta=3$, the number of hidden layer units being 650, and total vocabulary size to be 10k.

Table \ref{tab:jumble} shows the result.
While all the variants of \jlm achieve high accuracy, the statistical test revealed that INT and END have a difference with statistically marginal significance ($p=0.07$).
There are statistically significant differences ($p<0.01$) both in END\&BEG and BEG\&ALL.
From the results, the word recognition difficulty of different jumbled types is summarized as INT $\leq$ END $<$ BEG $<$ ALL, which is the same order as the finding in Table \ref{tab:previous}. 
It is not surprising that INT outperform the other variants because it has richer representation in $x_n$ (twice or three times larger than the other variants). 
However, it is interesting to see that END outperforms BEG both in \newcite{Rayner01032006} and our experiment despite that the size of $x_n$ between END and BEG models are equal.
This suggests the \jlm replicates (at least a part of) the human word recognition mechanism, in which the first letter is more important and informative than the last one in English.

For qualitative analysis, Table \ref{tab:error} shows some errors (correct/wrong) that each variant made.
All the \jlm variants often fail to recognize capitalized first character (e.g. Once/once, Under/under), specifically when the word is at the beginning of the sentence.
Other than the capitalization errors, most errors come from anagrams.
For example, errors in INT (the original \jlm) are internally anagrammable words (e.g. there/three, form/from).
END model made errors on words that are anagrammable with the first character being fixed (e.g. being/begin, quiet/quite). 
BEG model, on the other hand, failed to recognize anagrammable words with the last character being the same (e.g. creditors vs. directors, views vs. wives). 
In addition, BEG model often ignores the first (upper-cased) character of the word (e.g. Several/reveal, Growth/worth).
Finally, ALL model failed to recognize anagrammable words (e.g. center/recent, licensed vs. declines).
Although \jlm generally disambiguated anagrammable words successfully from context, all these examples from the error analysis are straightforward and convincing when we consider the characteristics of each variant of \jlm.


\section{Summary}
We have presented a semi-character recurrent neural network model, \jlm,  which is inspired by the robust word recognition mechanism known in psycholinguistics literature as the \cambridge effect.
Despite the model's simplicity compared to character-based convolutional neural networks (\cnn), it significantly outperforms widely used spelling checkers with respect to various noise types.
We also have demonstrated a similarity between \jlm and human word recognition mechanisms, by showing that \jlm replicates a psycholinguistics experiment about word recognition difficulty in terms of the position of jumbled characters.

There are a variety of potential NLP applications for \jlm where robustness plays an important role, such as normalizing social media text (e.g. {\em Cooooolll $\rightarrow$ Cool}), post-processing of OCR text, and modeling morphologically rich languages, which could be explored with this model in future work.


\bibliographystyle{aaai}
\bibliography{aaai}

\begin{thebibliography}{}

\bibitem[\protect\citeauthoryear{Ballesteros, Dyer, and
  Smith}{2015}]{ballesteros-dyer-smith:2015:EMNLP}
Ballesteros, M.; Dyer, C.; and Smith, N.~A.
\newblock 2015.
\newblock Improved transition-based parsing by modeling characters instead of
  words with lstms.
\newblock In {\em Proceedings of the 2015 Conference on Empirical Methods in
  Natural Language Processing},  349--359.
\newblock Lisbon, Portugal: Association for Computational Linguistics.

\bibitem[\protect\citeauthoryear{Bengio, Simard, and
  Frasconi}{1994}]{bengio1994learning}
Bengio, Y.; Simard, P.; and Frasconi, P.
\newblock 1994.
\newblock Learning long-term dependencies with gradient descent is difficult.
\newblock {\em IEEE transactions on neural networks} 5(2):157--166.

\bibitem[\protect\citeauthoryear{Chrupala}{2013}]{DBLP:journals/corr/Chrupala13}
Chrupala, G.
\newblock 2013.
\newblock Text segmentation with character-level text embeddings.
\newblock {\em arXiv preprint arXiv:1309.4628}.

\bibitem[\protect\citeauthoryear{Chrupa{\l}a}{2014}]{chrupala:2014:P14-2}
Chrupa{\l}a, G.
\newblock 2014.
\newblock Normalizing tweets with edit scripts and recurrent neural embeddings.
\newblock In {\em Proceedings of the 52nd Annual Meeting of the Association for
  Computational Linguistics (Volume 2: Short Papers)},  680--686.
\newblock Baltimore, Maryland: Association for Computational Linguistics.

\bibitem[\protect\citeauthoryear{Church, Hart, and
  Gao}{2007}]{church-hart-gao:2007:EMNLP-CoNLL2007}
Church, K.; Hart, T.; and Gao, J.
\newblock 2007.
\newblock Compressing trigram language models with {Golomb} coding.
\newblock In {\em Proceedings of the 2007 Joint Conference on Empirical Methods
  in Natural Language Processing and Computational Natural Language Learning
  (EMNLP-CoNLL)},  199--207.
\newblock Prague, Czech Republic: Association for Computational Linguistics.

\bibitem[\protect\citeauthoryear{Davis}{2003}]{davis2003aoccdrnig}
Davis, M.
\newblock 2003.
\newblock Aoccdrnig to a rscheearch at {C}mabrigde {U}inervtisy.
\newblock {\em http://www.mrc-cbu.cam.ac.uk/people/matt.davis/cmabridge/}.

\bibitem[\protect\citeauthoryear{Forster \bgroup et al\mbox.\egroup
  }{1987}]{forster1987masked}
Forster, K.~I.; Davis, C.; Schoknecht, C.; and Carter, R.
\newblock 1987.
\newblock Masked priming with graphemically related forms: Repetition or
  partial activation?
\newblock {\em The Quarterly Journal of Experimental Psychology}
  39(2):211--251.

\bibitem[\protect\citeauthoryear{Grainger \bgroup et al\mbox.\egroup
  }{2006}]{grainger2006letter}
Grainger, J.; Granier, J.-P.; Farioli, F.; Van~Assche, E.; and van Heuven,
  W.~J.
\newblock 2006.
\newblock Letter position information and printed word perception: the
  relative-position priming constraint.
\newblock {\em Journal of Experimental Psychology: Human Perception and
  Performance} 32(4):865.

\bibitem[\protect\citeauthoryear{Guerrera and
  Forster}{2008}]{doi:10.1080/01690960701579722}
Guerrera, C., and Forster, K.
\newblock 2008.
\newblock Masked form priming with extreme transposition.
\newblock {\em Language and Cognitive Processes} 23(1):117--142.

\bibitem[\protect\citeauthoryear{Hochreiter and
  Schmidhuber}{1997}]{hochreiter1997long}
Hochreiter, S., and Schmidhuber, J.
\newblock 1997.
\newblock Long short-term memory.
\newblock {\em Neural computation} 9(8):1735--1780.

\bibitem[\protect\citeauthoryear{Humphreys, Evett, and
  Quinlan}{1990}]{Humphreys1990517}
Humphreys, G.~W.; Evett, L.~J.; and Quinlan, P.~T.
\newblock 1990.
\newblock Orthographic processing in visual word identification.
\newblock {\em Cognitive Psychology} 22(4):517 -- 560.

\bibitem[\protect\citeauthoryear{Johnson, Perea, and
  Rayner}{2007}]{johnson2007transposed}
Johnson, R.~L.; Perea, M.; and Rayner, K.
\newblock 2007.
\newblock Transposed-letter effects in reading: Evidence from eye movements and
  parafoveal preview.
\newblock {\em Journal of Experimental Psychology: Human Perception and
  Performance} 33(1):209.

\bibitem[\protect\citeauthoryear{Kim \bgroup et al\mbox.\egroup
  }{2015}]{DBLP:journals/corr/KimJSR15}
Kim, Y.; Jernite, Y.; Sontag, D.; and Rush, A.~M.
\newblock 2015.
\newblock Character-aware neural language models.
\newblock {\em arXiv preprint arXiv:1508.06615}.

\bibitem[\protect\citeauthoryear{Ling \bgroup et al\mbox.\egroup
  }{2015}]{DBLP:journals/corr/LingTDB15}
Ling, W.; Trancoso, I.; Dyer, C.; and Black, A.~W.
\newblock 2015.
\newblock Character-based neural machine translation.
\newblock {\em arXiv preprint arXiv:1511.04586}.

\bibitem[\protect\citeauthoryear{Mikolov \bgroup et al\mbox.\egroup
  }{2010}]{DBLP:conf/interspeech/MikolovKBCK10}
Mikolov, T.; Karafi{\'{a}}t, M.; Burget, L.; Cernock{\'{y}}, J.; and Khudanpur,
  S.
\newblock 2010.
\newblock Recurrent neural network based language model.
\newblock In {\em {INTERSPEECH} 2010, 11th Annual Conference of the
  International Speech Communication Association, Makuhari, Chiba, Japan,
  September 26-30, 2010},  1045--1048.

\bibitem[\protect\citeauthoryear{Perea and Lupker}{2004}]{Perea2004231}
Perea, M., and Lupker, S.~J.
\newblock 2004.
\newblock Can {CANISO} activate casino? transposed-letter similarity effects
  with nonadjacent letter positions.
\newblock {\em Journal of Memory and Language} 51(2):231 -- 246.

\bibitem[\protect\citeauthoryear{Peressotti and
  Grainger}{1999}]{Peressotti1999}
Peressotti, F., and Grainger, J.
\newblock 1999.
\newblock The role of letter identity and letter position in orthographic
  priming.
\newblock {\em Perception {\&} Psychophysics} 61(4):691--706.

\bibitem[\protect\citeauthoryear{Rayner \bgroup et al\mbox.\egroup
  }{2006}]{Rayner01032006}
Rayner, K.; White, S.~J.; Johnson, R.~L.; and Liversedge, S.~P.
\newblock 2006.
\newblock Raeding wrods with jubmled lettres: There is a cost.
\newblock {\em Psychological Science} 17(3):192--193.

\bibitem[\protect\citeauthoryear{Schmaltz \bgroup et al\mbox.\egroup
  }{2016}]{schmaltz-EtAl:2016:BEA11}
Schmaltz, A.; Kim, Y.; Rush, A.~M.; and Shieber, S.
\newblock 2016.
\newblock Sentence-level grammatical error identification as
  sequence-to-sequence correction.
\newblock In {\em Proceedings of the 11th Workshop on Innovative Use of NLP for
  Building Educational Applications},  242--251.
\newblock San Diego, CA: Association for Computational Linguistics.

\bibitem[\protect\citeauthoryear{Sutskever, Martens, and
  Hinton}{2011}]{sutskever2011generating}
Sutskever, I.; Martens, J.; and Hinton, G.~E.
\newblock 2011.
\newblock Generating text with recurrent neural networks.
\newblock In {\em Proceedings of the 28th International Conference on Machine
  Learning (ICML-11)},  1017--1024.

\bibitem[\protect\citeauthoryear{Van~Assche and Grainger}{2006}]{van2006study}
Van~Assche, E., and Grainger, J.
\newblock 2006.
\newblock A study of relative-position priming with superset primes.
\newblock {\em Journal of Experimental Psychology: Learning, Memory, and
  Cognition} 32(2):399.

\end{thebibliography}

\end{document}